\documentclass[letterpaper, 10 pt, conference]{ieeeconf}

\IEEEoverridecommandlockouts
\usepackage{amsmath} \usepackage{amssymb}  \usepackage{booktabs}

\usepackage{algorithm}

\usepackage{hyperref}
\usepackage{graphicx} \graphicspath{ {figures/} }
\newcommand{\circled}[1]{\raisebox{.5pt}{\textcircled{\raisebox{-.9pt} {#1}}}}

\newlength{\myfigwidth}
 \newlength{\myfigwidthhalf}
\usepackage[dvipsnames]{xcolor}

\newcommand{\yes}{\textcolor{ForestGreen}{$\checkmark$}}
\newcommand{\no}{\textcolor{red}{$\times$}}

\title{\LARGE \bf
The ACRV Picking Benchmark:\\% --\\%
A Robotic Shelf Picking Benchmark to Foster Reproducible Research
}

\author{J\"urgen Leitner$^{1,2}$,           Adam W.~Tow$^{1,2}$,                Niko S\"underhauf$^{1,2}$,          Jake E.~Dean$^{2}$,                 Joseph W.~Durham$^{3}$,             Matthew Cooper$^{2}$,\\          Markus Eich$^{1,2}$,                Christopher Lehnert$^{2}$,          Ruben Mangels$^{2}$,                Christopher McCool$^{2}$,           Peter~T.~Kujala$^{1,2}$,            Lachlan\\Nicholson$^{2}$,           Trung Pham$^{1,4}$,                 James Sergeant$^{1,2}$,             Liao Wu$^{2}$,                      Fangyi Zhang$^{1,2}$,               Ben Upcroft$^{1,2}$,                and Peter Corke$^{1,2}$\thanks{This research was supported by the Australian Research Council Centre of Excellence for Robotic Vision (ACRV) (project number CE140100016). The participation at the APC was supported by Amazon Robotics LLC.}\thanks{$^{1}$Authors are with the Australian Centre for Robotic Vision (ACRV)}\thanks{$^{2}$Authors are with the Queensland University of Technology (QUT), Brisbane, QLD 4001 Australia.}\thanks{$^{3}$JWD is with Amazon Robotics LLC, North Reading, MA 01864 USA.}.\thanks{$^{4}$TP is with the University of Adelaide, Adelaide, SA 5005 Australia.}\thanks{{\tt\small j.leitner@roboticvision.org}}}

\begin{document}

\maketitle
\thispagestyle{empty}
\pagestyle{empty}

\begin{abstract}
Robotic challenges like the Amazon Picking Challenge (APC) or the DARPA Challenges are an established and important way to drive scientific progress. They make research comparable on a well-defined benchmark with equal test conditions for all participants. However, such challenge events occur only occasionally, are limited to a small number of contestants, and the test conditions are very difficult to replicate after the main event.
We present a new \emph{physical} benchmark challenge for robotic picking: the ACRV Picking Benchmark. Designed to be reproducible, it consists of a set of 42 common objects, a widely available shelf, and exact guidelines for object arrangement using stencils. A well-defined evaluation protocol enables the comparison of \emph{complete} robotic systems -- including perception and manipulation -- instead of sub-systems only.
Our paper also describes and reports results achieved by an open baseline system based on a Baxter robot.

\end{abstract}

\section{Introduction}

\begin{figure}[t!]
  \centering
  \includegraphics[width=\myfigwidth]{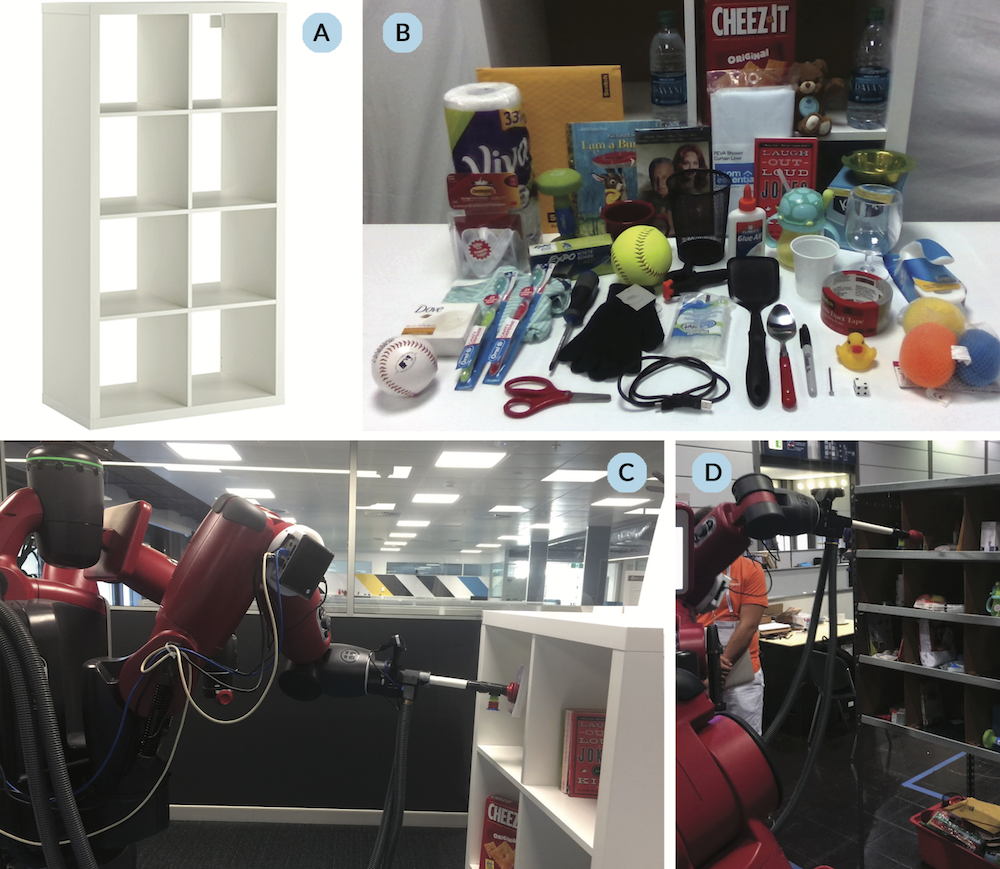}
   \vspace{-6.2mm}
\caption{The proposed benchmark consists of a commonly available shelf (A) and objects (B). A baseline system is presented performing the benchmark picking tasks (C) and during the Amazon Picking Challenge (D).}  \label{fig:vis-abstract}
  \vspace{-4.8mm}
\end{figure}

Robotic picking of randomly presented objects is one of the canonical problems in robotics. Applications range from household cleaning to space sample return, with considerable
economic potential in a wide range of industries, such as e-shopping and logistics. Despite a long history of research into dexterous manipulation~\cite{ulbrich2011opengrasp},
much of the observable recent progress has been driven by technological developments such as quality low-cost arms -- e.g.\ Rethink's Baxter (as depicted in Fig.~\ref{fig:vis-abstract}D), Kinova's MICO and JACO arms~\cite{kinova}, etc.~-- and
simple compliant grippers based on coffee grounds~\cite{brown2010universal} or
suction~\cite{monkman2007robot,manipulators,APCLessonslearned}.  This in turn has revealed the deficiencies in state-of-the-art hand-eye coordination and robust perception in real-world environments.

Over more than a decade the robotics community has embraced challenges as a means to drive progress -- from self-driving cars, to humanoids, to robotic picking. The methodology of using standard problems is powerful but the challenge events occur only occasionally and the test conditions are difficult to replicate outside the event.
Alongside challenges, there is growing interest in reproducible research~\cite{ReplicableRAM} which is relevant and related.

In parallel to activity within the robotics community, the vision research community has made enormous progress in problems such as object recognition and detection using dataset challenges such as the ImageNet-based ILSVRC~\cite{ILSVRC15}, COCO~\cite{Coco}, and Pascal-VOC~\cite{PascalVOC}.
However, pure image dataset challenges are of limited use for robotics where the captured images can vary widely.

This paper is motivated by our experience in the 2016 Amazon Picking Challenge (APC).  The challenge is a very effective way to drive progress, but it occurs only annually and is limited to 16 participant teams who are provided with the appropriate physical artifacts: standard shelf, standard set of objects. To drive progress we believe that it is essential to make the challenge conditions more widely available.
We propose a benchmark, based on commonly available shelves and objects, to allow for an easier reproduction of robotic picking, hopefully
allowing for more thorough analysis and better comparison.  In addition, we describe an open-source baseline system for this shelf picking benchmark. Our system is based on a Baxter robot (depicted picking from two shelves in Fig.~\ref{fig:vis-abstract}) and includes extensions which we made publicly available to enable replication. We report our systems performance on four exemplary setups of our benchmark.

\begin{table*}[bt]
    \caption{Comparison of the proposed benchmark with existing datasets and benchmarks.}
    \label{tbl:benchmarks}
    \begin{center}
        \vspace{-3mm}
        \begin{tabular}{@{}lccccc@{}}
        \toprule
        & \cite{ILSVRC15, Coco} & OpenGrasp \cite{ulbrich2011opengrasp} & YCB \cite{Calli2015} & Amazon Picking & \textbf{ACRV Picking}  \\
        & \cite{PascalVOC, singh2014bigbird} & VisGraB \cite{Popovic2011} & & Challenge \cite{APCLessonslearned} & \textbf{Benchmark (ours)} \\
        \midrule
        Perception                      & \yes  & \no   & \yes & \yes  & \yes \\
        Manipulation                    & \no   & \yes  & \yes & \yes  & \yes \\
        Real-world performance          & \no   & \no   & \yes & \yes  & \yes \\
        Cluttered / confined workspace  & \no   & \no   &  \no & \yes  & \yes \\
        \midrule
        Objects available               & \no   & \no   & \yes & \yes  & \yes \\
        Evaluation protocol             & \yes  & \yes  & \yes & \yes  & \yes \\
        Easily reproducible             & \yes  & \yes  & \yes & \no   & \yes \\
        Multiple degrees of difficulty  & \no   & \no   & \no  & \no   & \yes \\
        \bottomrule
        \end{tabular}
        \vspace{-4mm}
    \end{center}
\end{table*}

\section{Related Work -- Benchmark Datasets and Challenges in Robotic Manipulation Research}

Widely accepted benchmark datasets and challenges foster reproducible research and can drive scientific progress. The recent progress in computer vision for tasks like object detection, scene labeling, or visual question answering has been largely stimulated by datasets and challenges like ILSVRC~\cite{ILSVRC15}, COCO~\cite{Coco}, or Pascal-VOC~\cite{PascalVOC}.

Robotics manipulation in general, but especially pick-and-place of generic everyday objects in cluttered environments, is a highly complex task that comprises perception, path planning and execution, as well as grasping. Posing a benchmark that measures the performance of a robotic system for the overall task is challenging. Related work in this area often concentrated on solving only one of the sub-problems. Therefore a number of benchmarks and datasets exist that focus only on perception, grasping, or path planning.

\subsection{Perception-Only Datasets and Benchmarks}
Apart from the perception datasets established by the computer vision community~\cite{ILSVRC15,PascalVOC,Coco}, related work proposed datasets and benchmarks that focus on object perception in a robotics setting. The BigBIRD dataset~\cite{singh2014bigbird} provides images, point clouds, and meta data for 125 objects, obtained on a turn table.
\cite{tejani2014latent} introduced a 2D/3D database for multiple-instance detection with considerable clutter and foreground occlusion.
The increasing number of RGB-D datasets, e.g.\ for object segmentation~\cite{lai2013rgb} and 6D pose estimation~\cite{rennie2016dataset} can be contributed to the ubiquity of low-cost depth sensors.

\subsection{Manipulation-Only Datasets and Benchmarks}

Manipulation is an important problem in robotics and various benchmarks have been proposed to benchmark the state of the art capabilities of robotic systems.
VisGraB\cite{Popovic2011} and OpenGRASP \cite{ulbrich2011opengrasp} provide open frameworks to compare object manipulation capabilities. While these focus more on simulation, the YCB \cite{Calli2015} dataset defines a set of (physical) objects that cover a wide range of aspects of the manipulation problem, including size, weights and rigidity.
The focus of these datasets though lies mostly in the grasping aspects, removing perception and planning complexities.

\subsection{Complete-System Benchmarks}
The robotics community has embraced challenges to drive progress and benchmark complete system performance.
The RoboCup@Home~\cite{Wisspeintner2009} competitions, for example, are held annually and cover a wide area of robotic system technologies with a focus on domestic use robots. One part of the challenge is object detection and manipulation in a full room setup. The changing environment is built up under substantial effort every year. Additionally, teams are allowed to decide some of the task specifics, e.g.\ which objects to pick, making it even harder to compare performance across events.

The Amazon Picking Challenge (APC) \cite{APCLessonslearned} is the most recent attempt to create a benchmark for robotic picking and stowing that measures a system's performance for the \emph{complete} task. Its major drawback is the lack of reproducibility: while Amazon provided all participants with objects and the proprietary Kiva shelf used during the competition, only 16 teams were accepted into the challenge. Lacking the shelf, the objects, instructions how to place the objects in the shelf, and the scoring details, it is not possible to reproduce the benchmark. Furthermore, the best teams scored almost perfectly, indicating that the posed task is solved given the state of the art.
The YCB dataset also includes task specifications, yet these setups lack cluttered and confined environments, such as commonly seen during APC and RoboCup@Home competitions.

\subsection{Towards the Ideal Robotic Picking Benchmark}

\noindent The ideal robotic picking benchmark should:   \begin{itemize}
    \item measure the performance of the overall system, including perception, grasp planning, motion planning, and execution, instead of being limited to a subset;
    \item evaluate real-world performance instead of relying on simulation;
    \item evaluate the performance in a non-idealised environment, e.g.\ cluttered or confined spaces, that are to be expected in many industrial and domestic applications;
    \item provide access to all required components and physical artifacts, such as objects and shelf;
    \item provide an evaluation protocol enabling meaningful performance comparison of different robot systems;
    \item be easily reproducible by participants by clear instructions and well-defined setups;
    \item comprise several levels of difficulty with a low entry barrier for new researchers in the field, but also challenging scenarios that go beyond the state of the art in order to drive research forward.
\end{itemize}

Table \ref{tbl:benchmarks} compares several existing benchmarks and datasets against these criteria. While the recent Amazon Picking Challenge and the YCB dataset come close to the ideal robotic picking benchmark, we identified three important missing features:  While YCB~\cite{Calli2015} proposes a number of very well-defined tasks, such as pouring liquid from a container, pick and place, or setting a table, these tasks are not required to be executed in cluttered or confined workspaces. On the other hand, the Amazon Picking Challenge requires a robotic system to operate in such a realistic and challenging scenario, but is not easily reproducible since it relies on the proprietary Kiva shelf and only accepts a few teams into the challenge every year. Both YCB and APC also lack multiple degrees of difficulty and scenarios that are beyond the current state of the art, and thus pose research problems to drive the field forward.
We therefore identified the need for a new picking benchmark that combines all these features and propose the \textit{ACRV Picking Benchmark}.

\section{The ACRV Picking Benchmark (APB)}
\label{sec:benchmark}

Given the limitations of current datasets we introduce a new \emph{physical} benchmark and evaluation protocol for robotic picking from shelving units inspired by the APC. The \textit{ACRV Picking Benchmark} (APB) is designed to foster reproducible and thus comparable results. It furthermore allows robotic systems to be evaluated on the \emph{complete} task, not just sub-tasks such as object detection, grasp selection, or path planning.
At this time the
APB highlights current research questions but it is also devised to accommodate more complex item arrangements and reasoning as progress is made.

A major focus through the design of this benchmark was to maximise reproducibility: a number of carefully chosen scenarios with detailed instructions on how to place, orient, and align objects with the help of printable stencils are defined.
To make the benchmark as accessible as possible to the research community, a white IKEA shelf (Fig.~\ref{fig:vis-abstract}A) is used for all picking tasks. Furthermore, we carefully curated a set of 42 objects (Fig.~\ref{fig:vis-abstract}B) to ensure global availability and reduced chance of import restrictions.\footnote{Food items in the YCB dataset and the rawhide dog chew toy could not be shipped to Australia for example and will violate import regulations in a number of other countries around the world as well.}

The defined benchmark scenarios vary in difficulty and challenge both the manipulation capabilities, and the perception pipeline, of the evaluated robotic system: the objects vary in weight and size, and comprise transparent, reflective, black and deformable objects. Some of the objects are prone to shifting their centre of mass when manipulated, and others must adhere to strong constraints on orientation, i.e.\ they are not allowed to be turned over. This creates challenges for the state of the art in robotic picking and hopefully yield improvements to our baseline system presented (Section \ref{sec:baseline}).

In the following, we describe selected aspects of the benchmark in greater detail.
The benchmark description, task definitions, object dataset (including labelled images), placement stencils (to print), and a 3D shelf model are available for download at: \url{http://Juxi.net/dataset/apb}

\subsection{The Shelf}
\begin{figure}[t]
  \centering
  \includegraphics[width=\columnwidth]{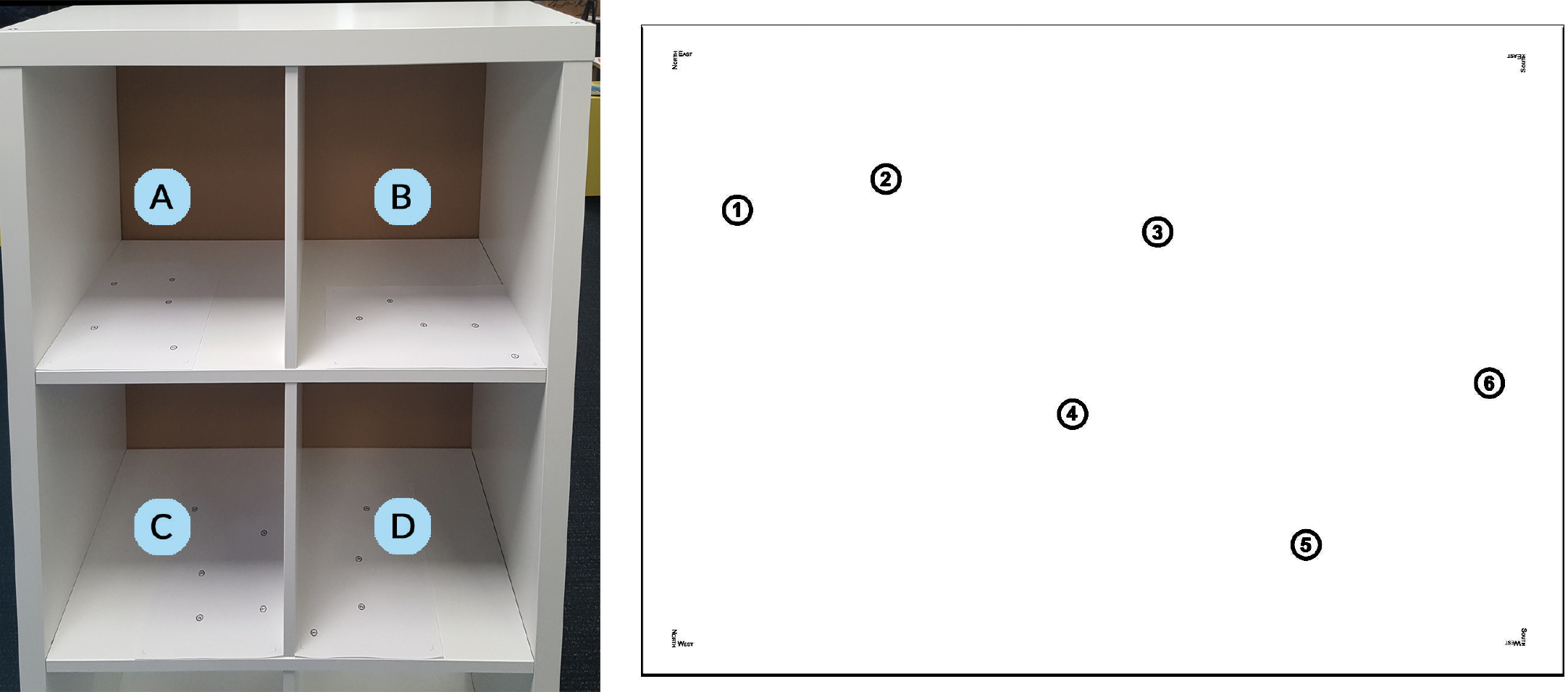}  \caption{(left) The upper 4 bins (labelled A to D) of the commonly available IKEA Kallax shelf are used for the benchmark.
    The back is covered with cardboard to ease the perception and resemble the conditions of the APC.
  (right) The \texttt{Dipper} stencil  with labelled markers for object placement.  }
  \label{fig:IKEAshelf}
  \vspace{-2mm}
\end{figure}

Our benchmark uses the commonly available white IKEA Kallax shelf with eight $33\times33\times39$ cm bins (w$\times$h$\times$d). The shelf can be ordered from IKEA\footnote{IKEA article number: 802.758.87, price: EUR 59/USD 65/AUD 75} and is conveniently available worldwide, unlike the proprietary Kiva shelf used during the APC.
For the tasks described in this paper, only the upper four bins are used and referred to as `Bin A' to `Bin D'.
To improve reproducibility and make the bins perceptually similar to the APC shelves, we cover the back of the shelf with cardboard from the packaging (Fig.~\ref{fig:IKEAshelf}).

\subsection{Object Dataset}
This benchmark consists of 42 unique objects (Fig.~\ref{fig:vis-abstract}B).
The items chosen present various challenges for both perception and manipulation. Each item is categorised into easy, medium, and hard difficulties for both detection and manipulability.
Item categorisations were selected based on our experience in the recent APC and with respect to our solution to the task. Each item was chosen to broaden the set of challenges presented. Some items, like the tissue box, are both easy to manipulate and relatively easy to detect. Other items, like the nail, are both difficult to detect (small, reflective) and difficult to manipulate. The set includes items similar in appearance to the background, items similar in appearance to others, and items invisible to low-cost depth sensors. Transparent, reflective, deformable and odd-shaped items are also included.
We also provide labelled image data taken with an Intel RealSense camera of all 42 objects in various configurations in the shelf.

\subsection{General Benchmark Task Description}
The benchmark evaluates the capabilities of a robotic system to identify and pick objects from a shelf and place them into a tote. It defines a number of \emph{setups} and calculates an evaluation score based on the robot's performance averaged over a number of these setups.
A \emph{setup} in the benchmark consists of two parts:
\begin{itemize}
    \item a \emph{layout} specification, defining the exact placement and alignment of objects in the shelf,
    \item a \emph{task description} (or \emph{work order}), defining which objects should be picked from the shelf
\end{itemize}

\begin{figure}[t]
  \centering
    \includegraphics[width=\myfigwidthhalf]{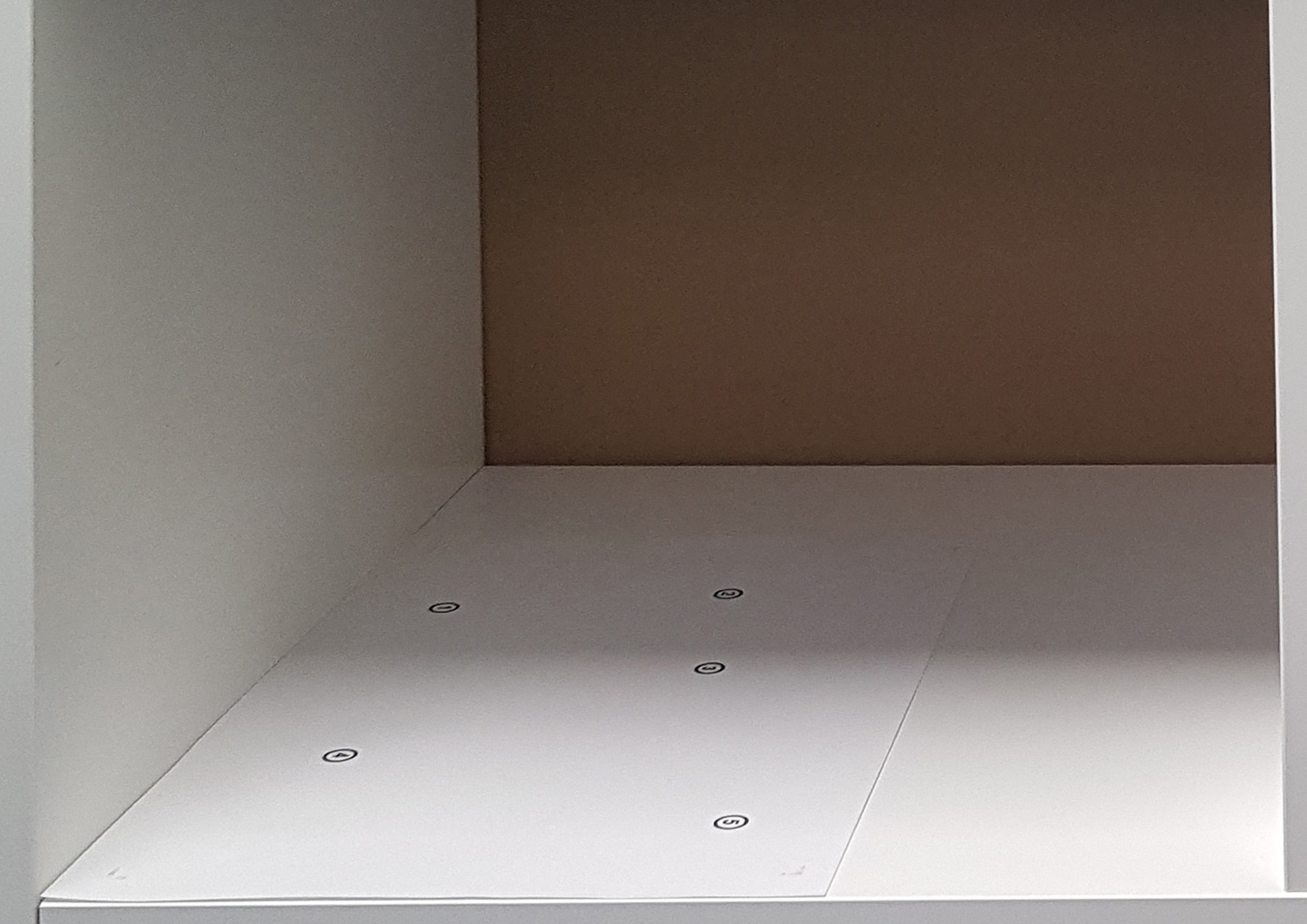}\hspace{0.2mm}\includegraphics[width=\myfigwidthhalf]{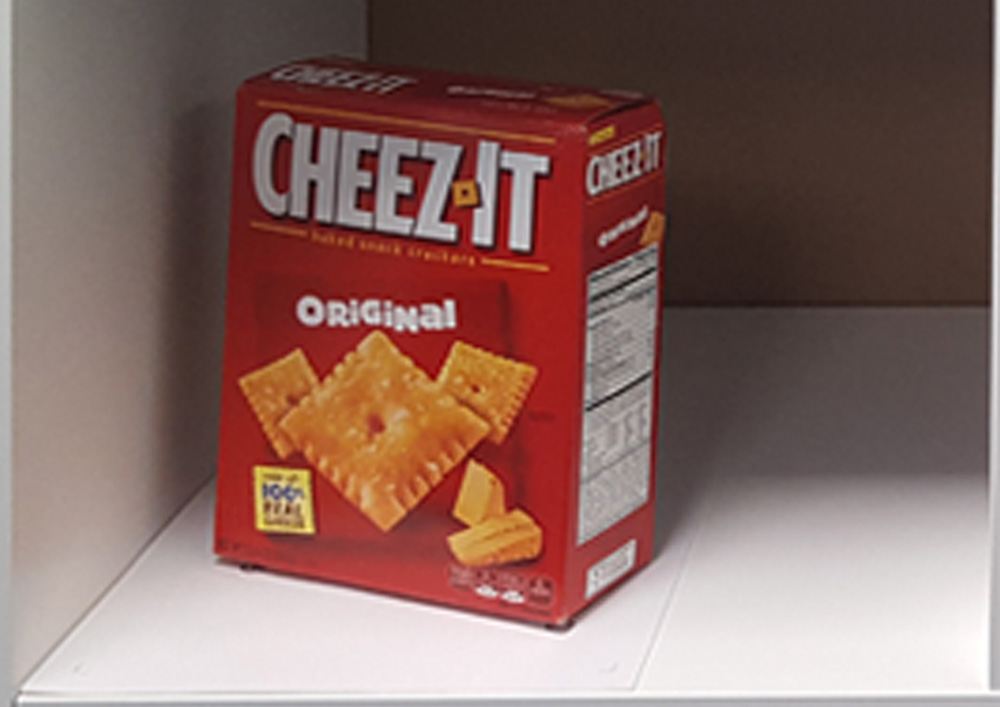}  \caption{    Placing objects with the help of stencil: (left) the \texttt{Crux} stencil is placed in the bin, then the object is placed on the markers (right)}
  \label{fig:StencilExample}
    \vspace{-5mm}
\end{figure}

\paragraph{Object Placement and Stencils}
To precisely control the placement of objects in the shelf bins, we created a set of \emph{stencils} (for both A4 and letter paper sizes). These feature light-grey, numbered markers (\circled{1}-\circled{9}) for objects to be placed on  according to a layout specification. For instance, a setup might call for the placement of object \texttt{cheezit} so that its front edge covers markers \circled{4} and \circled{5}, with the front corner of the object just covering marker \circled{5} (Fig.~\ref{fig:StencilExample}). A high-resolution photo of each setup will also be provided to remove ambiguity from the layout specification.
We aim to fully cover all markers of a stencil, so that object detection algorithms are not influenced by the stencil itself.

We provide 4 different, print-able stencils, inspired by and named after star constellations (e.g.\ \texttt{Cassiopeia}).
To place them, the layout description specifies how each stencil must be placed, each stencil
is marked with compass directions (N,E,S,W) to facilitate.
The four APB task setups presented require the stencils to be placed as follows:
\begin{itemize}
\item SW corner of \texttt{Crux} at the front, left corner of `Bin A';
\item NE \texttt{Cassiopeia}   at front, right of `Bin B';
\item NE \texttt{Crux}         at the front, right of `Bin C';
\item SE \texttt{Cassiopeia}   at the back, left of `Bin D'.
\end{itemize}

\begin{figure*}[tb]
  \centering
  \includegraphics[width=\linewidth]{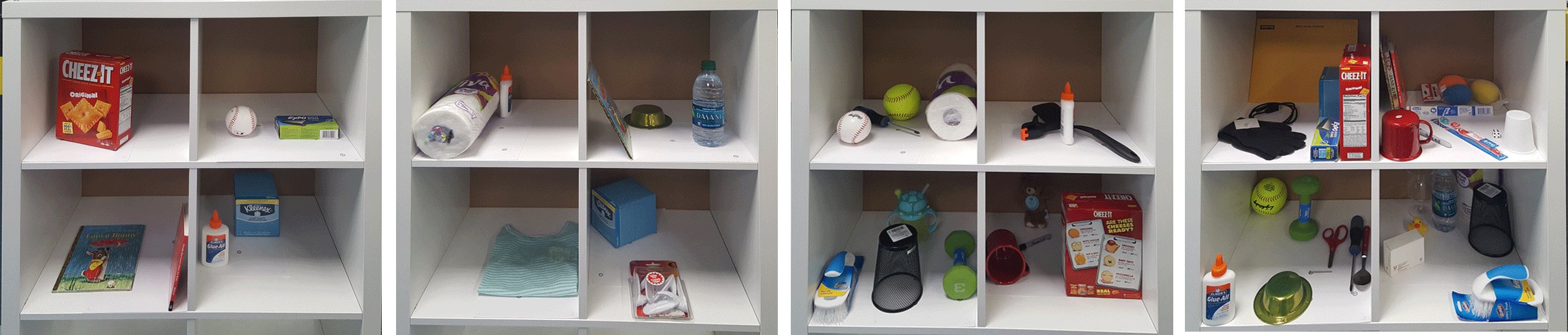}  \vspace{-1mm}
  \caption{Four picking setups, varying in their degree of difficulty from easy (left) to hard (right). In the hard setup a lot more objects are present. in addition for the hard setup on the very right, two objects need to be picked from two bins.}
  \label{fig:setups}
    \vspace{-4mm}
\end{figure*}

Images of each task setup can be seen in Figure~\ref{fig:setups}.
Pre-defined stencils allow precisely reproducible configurations of objects in the shelf. By providing multiple stencils a multitude of configurations can be created, limiting the risk of scripted solutions avoiding the perception problem -- furthermore the whole shelf is moved slightly between each run. More scenarios can be easily defined by the community in the future by combining or adding stencils.

\paragraph{Layout and Task Description}
A \textit{setup} is defined by a set of objects and stencils, and
includes a description of how objects are placed in the bins. The work order defines which objects are in each bin, and the objects to be picked and placed in the tote. This description is given to the robot in a JSON file (see Algorithm~\ref{JSON}), which will need to be continuously updated to represent the current state by the robot. Each bin has $n$ objects to pick ($n=1$, except in the hard task, where for two bins $n=2$).
A specific example of how such a setup is defined can be found in Section~\ref{sec:experiments}.
The object orientation can be determined from the image provided for each setup. The markers can also be used to specify precise orientation, e.g.\ centre object \texttt{sharpie\_marker} over \circled{2} with the tip pointing to \circled{3}.

\begin{algorithm}[tb]
  \caption{\tiny A JSON file snippet describing the state of the shelf and the work order for the first setup.}\label{JSON}
\scriptsize
\begin{verbatim}

{
  "bin_contents":
    { "bin_A": [ "cheeze-it_388g" ],
      "bin_B": [ "rawlings_baseball", ... ],
      ...                                       },
  "tote_contents": [],
  "work_order":
    [   {   "bin": "bin_A",
            "item": "cheeze-it_388g"  }, ...    ]
}
\end{verbatim}\normalsize
\end{algorithm}

\paragraph{Evaluation Score}
A simple success rate is used to evaluate the overall system performance.
A successfully picked object is an item that is in the work order and placed in the tote without being dropped from higher than 35cm and has not accrued damage during manipulation.
To ensure a certain robustness and reduce cherry-picking of results, the \emph{system performance} score is reported as the percentage of successful picks from the list of objects-to-pick, averaged over \textbf{three consecutive} runs.
The time for each successful pick shall be reported. The robot may not take more than 15 minutes to fulfil the full work order.

\subsection{Evaluation Guidelines}
We will curate a list of submissions and a leaderboard, where teams can submit scores (and new tasks). A video to verify the robot's score and show the robot in operation will be required. The recording allows the verification of the results, and the adding of scoring metrics retrospectively.  In addition, the teams can provide links to systems descriptions, publications, and code.
The following guidelines are proposed for fair comparing and ranking of submissions.\footnote{Details are available online: \url{http://Juxi.net/dataset/apb}}

The robot is not to be touched or tele-operated during the entire setup and scoring.
Each separate run begins with the setup phase. First the shelf is moved slightly (each corner's position can change within a 2cm square).
By moving, we hope to foster development of more generic, more robust perception solutions (and limiting  scripted solutions).
The stencils are then placed in the shelf bins before aligning the objects with the correct makers.
An image is provided with the task description for verification of each object's rotation and placement.
The tote where the objects will be placed can be positioned manually anywhere within a 2m workspace in front of the shelf. The tote is not allowed to touch the shelf or be rigidly fixed to the ground.
Any part of the robot shall not be closer than $0.5$m to any part of the shelf at the start.
After the shelf and objects are setup the robot is turned into autonomous operation and the clock is started.
As there is no pre-defined order, the system can choose in which order to pick.
A run is over when the work order is fulfilled (all items are picked) or the maximum time has elapsed.
The results for 3 consecutive runs shall be reported.

\section{A Baseline System for Shelf Picking}

\begin{figure}[t!]
  \centering
  \includegraphics[width=\myfigwidth]{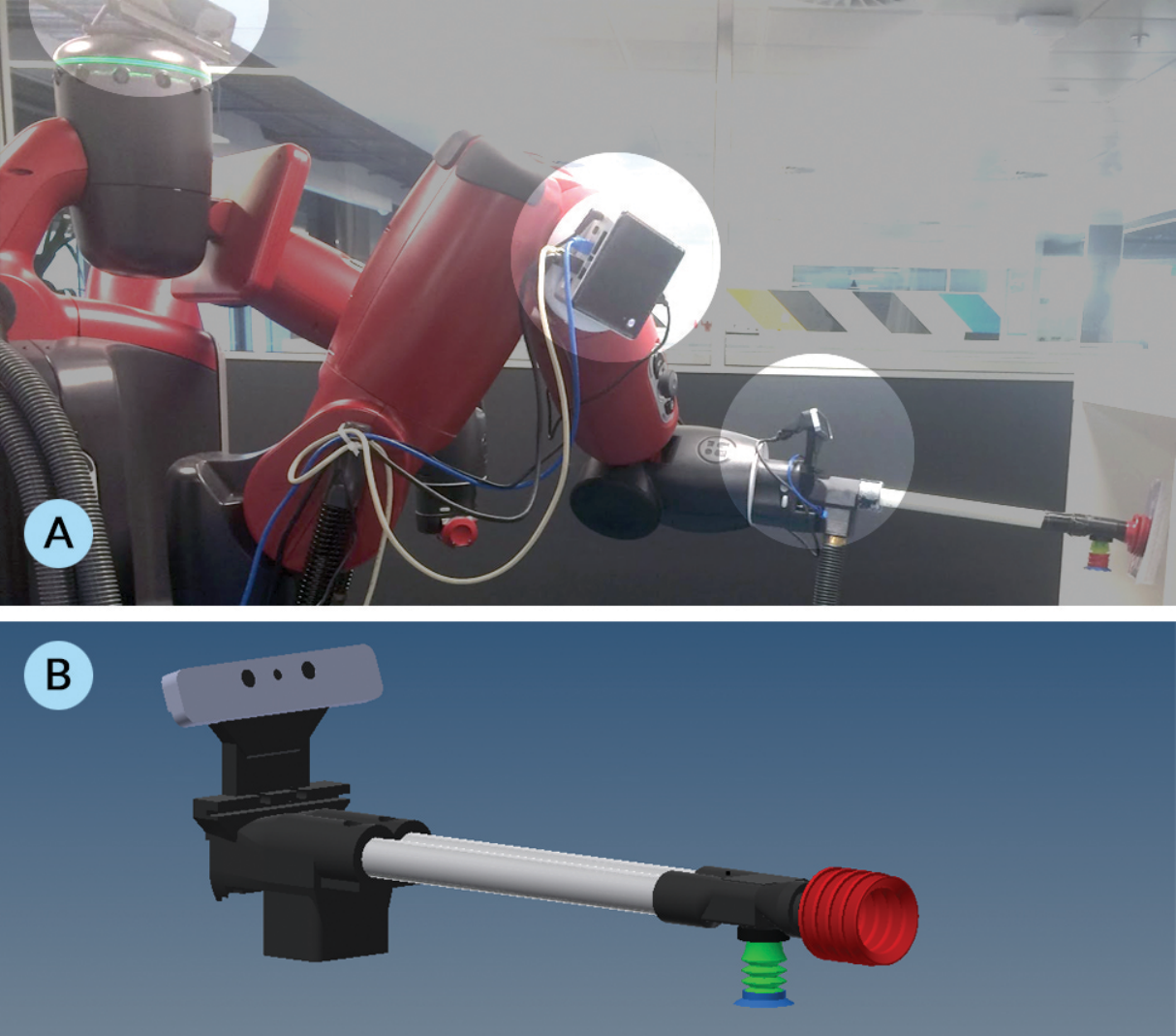}
  \vspace{-6mm}
  \caption{(A) The Baxter baseline system with highlights on the sensing and computing additions (a Kinect2, a NUC computer and an Intel Realsense SR300 camera). (B) End-effector design to reach into the Kiva shelf bins.  }
  \label{fig:setup}
  \vspace{-7mm}
\end{figure}

We present a benchmark shelf picking system primarily composed of inexpensive, readily available components. The platform leverages a single seven degree of freedom arm of a static Baxter robot.
To promote comparison against a variety of systems, a single-arm setup was chosen.

\begin{figure*}[t!]
  \centering
  \includegraphics[width=\linewidth]{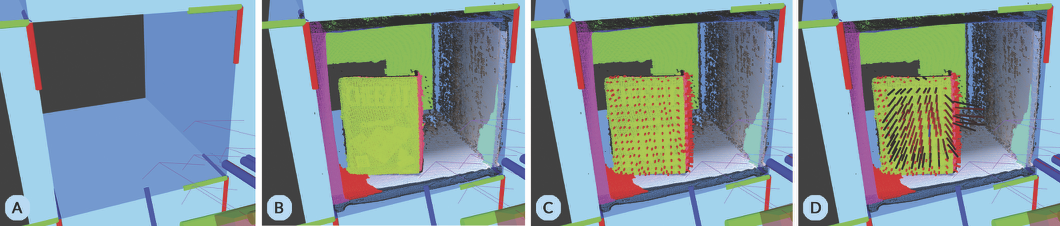}
  \vspace{-6mm}
  \caption{Our perception pipeline performs a shelf/bin localisation (A), followed by a segmentation of the fused point cloud (B) before proposing possible grasp candidates (in red) (C). A grasp point is then chosen (blue) based on a ranking heuristic, visualised by the length of the arrow (D).}
  \label{fig:perception}
  \vspace{-4mm}
\end{figure*}

\subsection{Baseline Approach}
The baseline system extends Baxter with additional perception capabilities, a custom end-effector with two suction cups at right angles, a small kinetic vacuum pump, and an Intel NUC PC mounted on the robot's elbow.

First the robot localises the shelf by using the Kinect2.
The robot then chooses the next object to pick from the work order provided (in a JSON file) and moves the robot's end effector to a pre-recorded scan pose. During a diamond-shaped scanning operation (parallel to the shelf's front), point clouds, provided by an Intel RealSense on the wrist (Fig.~\ref{fig:setup}), are recorded. A fused cloud is then sent into the perception pipeline, which segments and identifies the target object.
The grasping subsystem detects grasp points (and which of the two suction cups to use) in the provided segmented point cloud.
Following grasp point selection, the robot turns on the vacuum pump and plans and attempts to execute motions to pre-grasp and grasp poses. Successful object attachment is detected with an in-line pressure sensor. When a successful grasp is detected, trajectories to remove the object from the shelf and position the item over the tote are performed. When over the tote, the item is released by switching the vacuum pump off and the robot chooses the next item to attempt. If a grasp is unsuccessful, the robot returns to scan the scene to accommodate for times when items have moved within the bin. After three failed grasp attempts, an item is skipped.

Computation was provided by multiple computers running the Robot Operating System (ROS). Alongside the computer within Baxter, the Intel NUC mounted on Baxter's elbow drives the additional on-board cameras, provides point clouds over the network and reads the pressure sensor switch to detect successful grasps. An off-board computer with an NVIDIA TitanX GPU was used for perception, motion planning and to run the state machine.

\subsection{End-effector Design}

The robot's custom-designed gripper provides a mounting point for two suction cups, two vacuum lines and a small RGB-D camera. The design is released openly (under Creative Commons BY-SA 3.0 license) and CAD models are available for download from the benchmark webpage. The gripper consists of two 3D-printed components -- one connected to the wrist of the robot, the second to house the two suction cups -- connected by two parallel PVC pipes of 27cm length. Existing gripper mount points allowed for our gripper to be attached to Baxter non-destructively.
It was designed with the following features and considerations:

\begin{itemize}
    \item Small cross-section for reaching into filled bins
    \item Two suction cups mounted at right-angles for grasping objects frontward, downward and sideward
    \item Camera mount to maximise closeness to shelf front whilst minimising the joint-space distance between look-into and grasp-ready poses
    \item Mechanically simple design that was easy to manufacture quickly and optimise iteratively \end{itemize}
Suction cups provide a simple and effective way to grasp a wide variety of objects. Our design includes two separate cups to increase the number of reachable grasp points. Considering an object-filled shelf, the additional right-angled suction cup enables object grasps from above, as well as, side ways picks of objects leaning against the wall, especially books and DVDs. Furthermore, aligning the suction cup with the axis of rotation of the wrist creates the capability to grasp objects from the side (Fig.~\ref{fig:setup}B).
An RGB-D camera, mounted at a fixed-angle location, provides local shelf sensing. The angle chosen was empirically found by maximising closeness to the front face of the shelf and reducing the joint-angle distance required to reposition between look-into-shelf and pre-grasp pose.

\subsection{Perception Pipeline}

\subsubsection{Shelf Localisation}

The system leverages a head-mounted depth sensor (a Kinect 2) to localise the shelf wrt.\ the robot.
A single point cloud of the empty shelf was captured and the four front corners annotated by a human. These were then used to extract the grid structure of the shelf's front face.
A one-off, semi-autonomous pre-alignment is finalised using ICP~\cite{besl1992method} registration between the grid points and a 3D model of the shelf.

The autonomous alignment stage registers the pre-recorded point cloud of the shelf wrt.\ to the current point cloud provided by the sensor by NDT~\cite{magnusson2009three}.
Combining this with the transform found in the pre-alignment step a 3D model of the shelf is aligned to the live shelf position.
This approach is widely applicable to varying shelf configurations though was primarily chosen to combat shelves with reflective floors (such as the APC shelves). In such cases, the difference between a Kinect cloud and a cloud generated from a 3D CAD model is large, necessitating some pre-processing to robustly align the points.

\subsubsection{Scan, Propose, Classify}

Our perception pipeline operates on the close-range data provided by the RealSense. The pipline was split into three distinct stages: a Kinect Fusion (KinFu)~\cite{newcombe2011kinectfusion} approach whilst running a scan motion, a point-cloud-based segmentation algorithm, and a classification stage using a deep convolutional network (CNN).

A point-cloud-based segmentation algorithm \cite{Pham2016} yields distinct object clouds.
These are projected into the current image frame to generate non-overlapping 2D object proposals, significantly reducing overall computation compared to EdgeBoxes \cite{Zitnick2014}. To further reduce the number of proposals by the chosen segmentation algorithm, we integrated a KinFu scanning motion, which creates a dense point cloud of the bin, leading to larger, more complete object segments. In addition it reduces the number of erroneous points from it.

Following the scan and segmentation object classification is performed.
At first a simple feature matching based on standard features, such as SIFT and SURF, was implemented.
Our modular design of the perception pipeline made swapping out classification algorithms quite easy.
Due to its superior performance over simple feature extractors a CNN was used to classify each object proposal.
The CNN was based on GoogLeNet~\cite{GoogLeNet} fine-tuned with 150 images per class (for 38 of the 39 object classes, as we were not able to import one object).

\subsection{Grasping Pipeline}
The objective of the grasping pipeline is to provide reachable grasp points on the object to pick. Provided a point cloud segment, candidate grasp points are generated by smoothing the cloud, estimating it's boundaries and computing point normals in a grid-like pattern across the surface. This step is fast to perform and includes hyper-parameters for the suction cup dimensions to reduce/increase the number of candidate points generated.

A grasp selection process follows. First, inverse kinematics for each candidate are checked at both pre-grasp (5 cm above) and grasp poses to ensure the robot can reach the object along the candidate point normal. Reachable grasps are then ranked with heuristics such as distance to the clouds boundary, curvature at the normal and distance to the walls of the bin. These heuristics were chosen to give our system the best chance of grasping an object in its current pose.

\subsection{Motion Planning}
Reaching is performed in open loop on an estimate of the real environment. Facilitated by a localised shelf model, RRT$^*$~\cite{karaman2011sampling} coupled with the TRAC-IK inverse kinematics solver was used to compute collision-free paths for moving the end-effector to the desired target pose. This combination of planner and IK-solver was empirically found to produce more reliable and consistent paths than the other planners and IK solvers tested. In order to optimise the workspace of Baxter's arm, Baxter was rotated roughly 45 degrees with respect to the shelf (see Fig.~\ref{fig:vis-abstract}C).

\subsection{Software}
We leverage the Robot Operating System (ROS) framework and additional open source software to speed up development of the system and promote modularity, standardisation and reproducibility. The baseline system's source code is publicly available at: \url{https://github.com/amazon-picking-challenge/team_acrv}.
Software packages integrated in the system include:
\begin{itemize}
    \item MoveIt! and OMPL - Motion Planning
    \item TRAC-IK - Inverse Kinematics Solver
    \item PCL - Point Cloud Library~\cite{Rusu2011}
    \item Iai\_kinect2 - Microsoft Kinect 2 Driver
    \item Librealsense - Intel RealSense SR300 Driver
    \item SMACH - State Machine
\end{itemize}

\section{Baseline Results}
\label{sec:baseline}
\label{sec:experiments}

The robotic shelf-picking system described above was initially developed for our participation in the Amazon Picking Challenge 2016 where we ranked 6th of 16 in the picking task. During discussions at the event the idea of a more generic benchmark proposed.
The following is an evalution of our APC system on the benchmark proposed in Section~\ref{sec:benchmark}. Only minor adjustments to the shelf registration and additional scripted way-points for out-of-shelf movements were added to transfer from APC to APB.

\subsection{ACRV Picking Benchmark}
The system was placed in front of the IKEA shelf, with the centre of Baxter's arm base $1.5$m away from the rear of the shelf and rotated by about $45^{\circ}$ to maximise shelf reachability.
The tote is manually placed at the same position for every run, between Baxter and the shelf, about 10cm away from the shelf. The vacuum pump and other equipment was attached to the back and/or the base of Baxter.

The baseline system was tested on four setups with increasing complexity (Fig.~\ref{fig:setups}).
For the `easz' task, our systems results were quite consistent, while for the more complicated tasks, our system was unable to robustly produce object segments and so resulted in a poor score.
Reflective or black objects, which are used in the hard task, create noisy read out from the depth sensor employed (coded light and near infrared projector) leading to ``holes'' in the point clouds. While we can mitigated to some extent by applying the Kinect Fusion motions, highly reflective or absorptive surfaces are problematic. The robust segmentation of the point clouds is crucial for high classification accuracy in our perception pipeline.

The `easy' task, represents the capabilities of current state of the art systems -- while there were teams that scored rather low or DNFs during the competition, the top scoring teams of the APC are likely to get perfect scores. The easy level is also included to emphasis the second metric used, \textit{quickest pick}.
In addition to our systems success rate over three runs in Table~\ref{tbl:APBBenchmark}, we report the quickest time-to-first-pick. This metric was chosen to foster research into faster robot systems.

\subsubsection*{Exemplary Task Description}
An example of a task description using the stencil placement as described above, is given for \textbf{Setup 1} below:
\begin{itemize}
\item `Bin~A': place object \texttt{cheeze-it\_388g} so that its front corner is over marker \circled{5}, its front edge just covers marker \circled{4} and \circled{5};
\item `Bin~B': place \texttt{rawlings\_baseball} centred over \circled{5};
\texttt{expo\_dry\_erase\_board\_eraser} front left corner over \circled{3}, front edge just covering \circled{2};
\item `Bin~C': place \texttt{i\_am\_a\_bunny\_book} front left corner over \circled{2} and left edge aligned with \circled{3};
\texttt{laugh\_out\_loud\_joke\_book} centred on \circled{1}, aligned with the front of the shelf, and leaning on the left wall;
\item `Bin~D': \texttt{elmers\_washable\_no\_run\_school\_glue} centred over \circled{1};
\texttt{kleenex\_tissue\_box} left front corner over \circled{3} and front edge parallel to the back wall.
\end{itemize}

On this setup our baseline system achieved a performance of $75\%$, ie.\ it picked 9 out of 12 objects, in three consecutive runs.
Our shelf localisation system was continuously running during all experiments, but
results presented do not include a movement of the shelf between the runs.

The setups are ordered by complexity. \textbf{Setup~2} adds complexity by introducing
deformable objects to the list of picks (\texttt{cherokee\_easy\_tee\_shirt} from `Bin C')
and occlusions (\texttt{platinum\_pets\_dog\_bowl} from `Bin B' and  \texttt{kleenex\_tissue\_box} from `Bin D'). In `Bin A' the \texttt{elmers\_washable\_no\_run\_school\_glue} is to be picked. Our first run needed to be aborted as the robot failed to execute any motion after starting. This result highlights the need to have multiple consecutive runs reported instead of single-shot, cherry-picked results. Overall we picked four objects (during the three runs).

The complexity of the picking task is further increased in \textbf{Setup 3}. Objects that are visually hard to differentiate are added, e.g.\ the object to pick from `Bin A' is \texttt{champion\_sports\_official\_softball}.
Also items that produce noisy point clouds were introduced, e.g.\ picking \texttt{plastic\_spatula} from `Bin B'.
In `Bin C' the target is \texttt{easter\_turtle\_sippy\_cup}, which has challenging geometry for manipulation.
Multi-object occlusions are also present throughout this setup (\texttt{cloud\_b\_plush\_bear} from `Bin D'), which focus not just on improved perception but also robust path planning. We were able to pick one item through the three runs (8.33\% success rate), due to segmentation problems (`holes' in point clouds due to black objects) and planning problems due to these inaccuracies.

\begin{table}[tb]
    \caption{Results of the baseline system on the benchmark tasks. Averaged over three consecutive runs.}
    \label{tbl:APBBenchmark}
    \begin{center}
        \vspace{-3mm}
        \begin{tabular}{@{}lrrr@{}}
        \toprule
        \textbf{Task}    &   \textbf{Success Rate} & \textbf{Successful Picks} & \textbf{Quickest Pick}\\
        \midrule
    Setup 1 &   $75.00\%$ & 9/12  (2/4, 4/4, 3/4) &  1:38 min \\      Setup 2 &   $33.33\%$ & 4/12  (0/4, 2/4, 2/4) &  3:08 min \\       Setup 3 &   $8.33\%$  & 1/12  (0/4, 1/4, 0/4) &  1:39 min\\
    Setup 4	&   $0.00\%$  & 0/18  (0/6, 0/6, 0/6) & no successful picks\\           \bottomrule
        \end{tabular}
        \vspace{-5mm}
    \end{center}
\end{table}

\textbf{Setup 4} is particularly hard. It aims to highlight the shortcomings of current systems and direction for robotics research.
The setup contains densely packed bins, very small objects, transparent objects and deformable objects.
The work order is to pick the deformable \texttt{usb\_cable\_1m} from `Bin A'.
In `Bin B' the two objects need to be picked, the \texttt{green toothbrush} which can not be detected as green in its current configuration, requiring some reasoning or verification after the pick.
Second the \texttt{jane\_eyre\_dvd} stacked between two books, requiring higher level planning.
In `Bin C' the single nail needs to be picked, as well as, the pair of scissors, which are visually very similar to the spoon placed just a couple of centimetres away.
Finally the ICRA duckie, which is small and of complex shape needs to be picked. Here our baseline system was not able to perform any grasps, due to the complexity of the perception, grasp detection, and the precise motion planning. Additionally our suction cup was not designed for tiny objects.

\addtolength{\textheight}{-1mm}

\section{Conclusions}
This paper is motivated by our experience in the recent Amazon Picking Challenge (APC) and discussions during the event.
Challenges, such as the APC, are effective in driving research but are sometimes hard to reproduce.
We propose a benchmark, with easily available physical artifacts: a standard shelf, standard set of objects, and reproducible task setups.
This benchmarking task  allows for more thorough analysis, better comparison and easier reproduction of \emph{complete} robotic picking tasks.

A major focus of the benchmark design was maximising reproducibility: a number of carefully chosen scenarios with precise instructions on how to place, orient, and align objects with the help of printable stencils are defined. In addition, a multitude of configurations can be created by combining the various stencils with all possible objects.
We carefully selected 42 objects that vary in weight and size, and include deformable, transparent, and closely related items (baseball and softball, red and green tooth brush, full and half full water bottle).

We see this work as the first phase and our hope is that the will expand on these setups. The  challenge will evolve as the competency of the community increases.
The defined benchmark scenarios vary in difficulty and challenge both the manipulation capabilities as well as the perception pipeline of the evaluated robotic system.
We are trying to strike a balance between reproducibility (to advance the science) and challenge (with immunity to gaming and cheating). Peer review by video, as used in some MOOCs, might be one avenue to explore, another might be auto-generated templates that are valid for a limited time (to counter scripting).

A system using a custom-off-the-shelf robot with publicly released hardware extensions is presented as a baseline.
It is representative of the state of the art -- it picked a score of $42$ points during the competition\footnote{\url{http://amazonpickingchallenge.org/results.shtml}} --
and is able to perform picks in the setups classified as `easy', while not being able to pick in `difficult' ones.
We hope the wider research community will take on this challenge and propose improvements to the baseline system and create novel solutions to improve robotic picking.

\bibliographystyle{IEEEtran}
\bibliography{IEEEabrv,IEEEexample,references,datasets}

\end{document}